\documentclass[letterpaper, 10 pt, conference]{ieeeconf}  

\IEEEoverridecommandlockouts                              

\usepackage{graphicx} 
\usepackage{hyperref}
\usepackage{cite}

\title{\LARGE \bf

We are all Individuals: The Role of Robot Personality and Human Traits in Trustworthy Interaction
}

\author{Mei Yii Lim$^{1}$, Jos\'e David Aguas Lopes$^{2}$, David A. Robb$^{1}$, Bruce W. Wilson$^{1}$, \\ Meriam Moujahid$^{1}$, Emanuele De Pellegrin$^{1}$ and Helen Hastie$^{1}$
\thanks{This work was funded and supported by the UKRI Node on Trust (EP/V026682/1).}
\thanks{$^{1}$School of Mathematical and Computer Sciences, Heriot-Watt University, EH14 4AS, UK
        {\tt\small m.lim@hw.ac.uk, d.a.robb@hw.ac.uk, bww1@hw.ac.uk, mm470@hw.ac.uk, 
        ed50@hw.ac.uk,
        h.hastie@hw.ac.uk}}%
\thanks{$^{2}$Semasio, Portugal
        {\tt\small jose@semasio.com}}%
}

\begin{document}

\maketitle
\thispagestyle{empty}
\pagestyle{empty}

\begin{abstract}
As robots take on roles in our society, it is important that their appearance, behaviour and personality are appropriate for the job they are given and are perceived favourably by the people with whom they interact. Here, we provide an extensive quantitative and qualitative study exploring robot personality but, importantly, with respect to individual human traits. Firstly, we show that we can accurately portray personality in a social robot, in terms of extroversion-introversion using vocal cues and linguistic features.  
Secondly, through garnering preferences and trust ratings for these different robot personalities, we establish that, for a Robo-Barista, an extrovert robot is preferred and trusted more than an introvert robot, regardless of the subject's own personality. Thirdly, we find that individual attitudes and predispositions towards robots do impact trust in the Robo-Baristas, and are therefore important considerations in addition to robot personality, roles and interaction context when designing any human-robot interaction study. 

\end{abstract}

\section{INTRODUCTION}
\label{sec:Introduction}
As robots become more capable of meaningful interaction, we will start to see them in settings that require social etiquette, such as in the service industry, including as baristas \cite{Lim22}, receptionists \cite{Meriam22} and bartenders \cite{Keizer14}. As these are everyday settings, they provide convenient platforms for experiments that do not require subjects to have specialist training, such as in the first responder or healthcare domains. 
Robots that serve in more of a social setting, such as service robots, as well as those used more as tools, such as surveying drones, all need to be trusted to do their job well \cite{Wynne18}. 
The Robo-Barista scenario discussed here, thus facilitates long term ``in the wild'' Human Robot Interaction (HRI) experimentation, allowing for the exploration of human-robot trust over time. Subsequently, this will, hopefully, encourage the adoption and acceptance of robots in our everyday lives.

Prior work has shown that both robot and human personality are significant predictors of trust \cite{Hancock21}. According to Joosse et al. \cite{Joosse13}, people implicitly assume that certain roles or occupations require certain personalities, but which personality would be appropriate for our Robo-Barista? When Howarth \cite{Howarth69} asked subjects to list occupations under extrovert and introvert categories, he found that service industry jobs such as waiter/ess, steward/ess and bartender were all listed under the extrovert categories. Furthermore, Sandstrom and Dunn \cite{Sandstrom14} found that people who interacted with a human barista as if they were a ``weak tie-friend", instead of a stranger, experienced more positive affect, mediated by a feeling of belonging. This led us to explore the research question of whether this finding maps over to HRI. Specifically, we explore here if a more extroverted Robo-Barista will be preferred and trusted over an introvert one. 

Trust in HRI is related to trust in automation and varying levels of trust in robots exist, which might result in misuse (over reliance in cases of extremely high levels of trust) or disuse of systems and robots (in cases of very low levels of trust) 
\cite{Lee04}. 
Furthermore, unfounded fluctuations in trust could inhibit user acceptance, thereby compromising the inherent advantages of technology \cite{Freedy07}. 
It is also important to take into consideration the person's attitude and propensity to trust, as well as the context of interaction and the task at hand \cite{Kraus20, Hancock21}.

As well as the commonly used `Negative Attitude to Robots' (NARS) measure \cite{Nomura06}, we also explore the subject's Propensity to trust (PTT), which is a relatively stable disposition that is developed over time. It can lead to trusting beliefs, i.e. perceived trustworthiness, which are formed as a trustor observes, interprets and ascribes motives to a trustee's actions \cite{Mayer95}, especially in novel situations \cite{McKnight98}. 
Rotter \cite{Rotter67} studied PTT in human-human interaction, while Jessup et al. \cite{Jessup19} explored PTT in human-automation interaction. Lewis et al. \cite{Lewis18} highlighted that there is very little empirical work that has investigated how users' PTT affects their relationship with a robot. In order to design robots for a specific context, continuing research on how, and to what extent, PTT plays a role in human-robot trust is necessary, hence it is one of the focuses of this paper.

\begin{figure}[thpb]
      \centering
      \framebox{
      \includegraphics[scale=0.25]{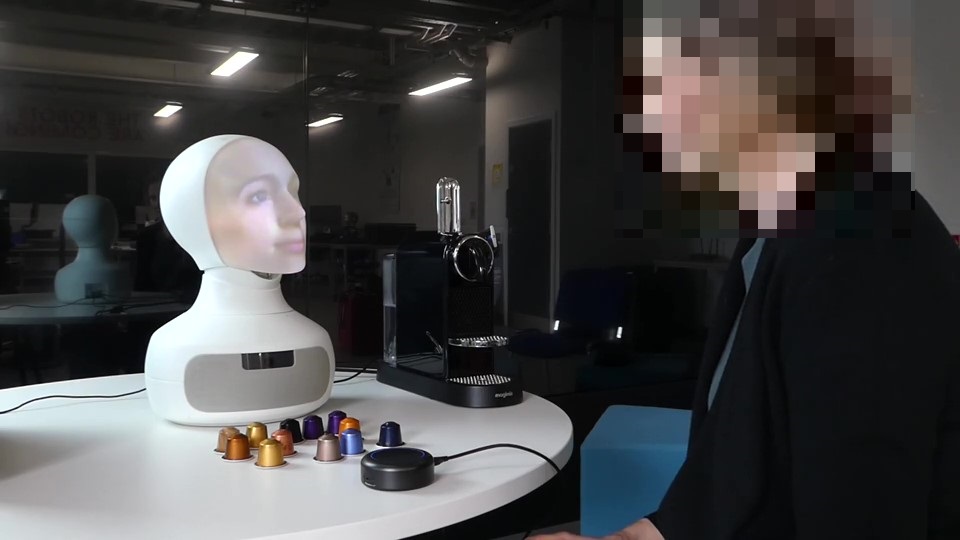}}
      \caption{The Robo-Barista interacting with a user}
      \label{fig:interactionsnapshot}
\end{figure}

As well as describing our general experimental platform for a Robo-Barista (Fig. \ref{fig:interactionsnapshot}), we address the following research questions: 

\begin{itemize}
    \item \textbf{R1:} Can extroversion and introversion be obviously implemented as personalities for a social robot, through linguistic and prosodic cues?
    \item \textbf{R2:} Which personality (extrovert or introvert) is preferred and trusted more for a service robot, such as a Robo-Barista? Do similarity/complementary-attraction theories between users and robots apply to a Robo-Barista? 
    \item \textbf{R3:} Are preference and trust measures correlated? 
    \item \textbf{R4:} Is trust in a service robot sensitive to individual differences and if so which ones? 
\end{itemize}

\section{Background}

While there is no generally agreed-upon definition of personality, it refers to traits that predict a person’s behaviour and one of the most used instruments is the Big Five theory, which proposes 5 personality types: extroversion, agreeableness, conscientiousness, neuroticism and openness to experience \cite{McCrae04}. It has been found to be robust and has external validity. Extraversion has been well-established to be positively related to dispositional interpersonal trust \cite{Gaines97, Kraus20} and was found to generalise to PTT in machines \cite{Merritt08, Evans08}. Much research on robot personality has been focused on the dimension of extroversion, as it is one of the easier traits to display, may be more salient in shorter interactions \cite{Robert18} and is, therefore, why we focused the robot's personality dimension along this axis.

Modulating prosody is an effective way to portray the extroversion-introversion dimensions in artificial agents \cite{Schmitz07, Tapus08} and has a strong influence on the perceived personality. 
In general, extroverts speak with louder voice, higher fundamental frequency, broader frequency range and faster speech rate than introverts \cite{Pittam94, Tusing00}. It is these prosodic parameters - volume, pitch and tempo - that we manipulate to create our two robot personalities of extrovert and introvert. Furthermore, there are pragmatic and lexical/syntactic differences \cite{Mairesse07, Neff10}. For example introverts use broader vocabulary, while extroverts are more repetitive and have fewer pauses and hesitations \cite{Furham90}.

Lohse et al. \cite{Lohse08} found significant preference for extrovert robot behaviour, which was rated more friendly, diversified, intelligent and faster. With regards perceived intelligence, Leuwerink's \cite{Leuwerink12} study showed that introvert robots are considered more intelligent in group interactions, while extrovert robots are more intelligent in dyadic interaction. Importantly, Alarcon et al. \cite{Alarcon18} demonstrate that individual differences play a role in the trust process. Studies have lent support for the idea that humans lean towards robots that have similar traits to them, an effect known as similarity-attraction \cite{Whittaker21}. However, this effect has not been unanimously demonstrated, with some studies showing there may be more of a complementary-attraction effect, that is, a preference for agents that complement their personality \cite{Lee06} but do not necessarily match it.  Joosse et al.'s work \cite{Joosse13} does align more with the complementary-attraction effect, 
but they posit that it depends on the interaction context and the robot role.  This adds further complexity to the personality matching debate.
Rousseau et al. \cite{Rousseau98} also stress the importance of context to understanding trust. Currently, there are not many studies on anthropomorphic robots in this area \cite{Milleounis15} and none in the context of a robot barista that we are aware of. This has led us to explore \textbf{R2} and \textbf{R3}.

With regards to trustworthy characteristics, Hancock et al. \cite{Hancock21} in their meta-review, found that robot-related, human-related and contextual factors are predictors of trust. Additionally, language indicators such as positive, gratitude and collaborative cues have been shown to help in identifying trustworthy interaction \cite{Lopes21}. Embodiment, social intelligence capabilities and non-verbal communication are other factors affecting trust \cite{Rheu21}. To answer \textbf{R4}, we will focus on human-related factors, specifically whether the subjects' Big 5 personality traits, PTT and NARS relate to perceived trustworthiness of the Robo-Barista.

\section{ROBO-BARISTA SET-UP}
\label{sec: Robo-Barista set-up}
We created a Robo-Barista to enable visitors and staff at our institution to talk and interact with a social robot, as illustrated in Fig. \ref{fig:interactionsnapshot}. The robot, takes drinks orders and interfaces with a coffee machine, which will automatically complete the order. The robot is able to detect when a potential customer is in the vicinity, either by the person directly addressing it or by it having detected their social presence. While it is preparing the drink, it engages the user in small talk and gives information on the institution. Finally, it alerts the user once the drink is ready and politely ends the conversation.

The Robo-Barista set-up consists of a Furhat robot from Furhat Robotics\footnote{\hyperlink{https://furhatrobotics.com}{https://furhatrobotics.com}} connected to a Nespresso CitiZ Espresso coffee machine. The dialogue was implemented using RASA 2.0\footnote{\hyperlink{https://rasa.com}{https://rasa.com}}. For Natural Language Understanding, we used their DIET model \cite{bunk2020diet} trained using data generated from template grammars. The DIET model handles both user intent classification and entity recognition in a single transformer architecture and allows for the use of pre-trained embedding models, such as BERT \cite{Bert19}. For the dialogue, we used the RASA TED Policy \cite{vlasov2019dialogue}, combined with a set of handwritten dialogue rules (e.g. for greetings and farewells). The TED Policy applies a machine learning policy that can generalise patterns from example conversations using context from previous utterances in the dialogue. The user may interject with off-topic remarks and have multiple embedded sub-dialogues (e.g., when the user changes their coffee order or engages in small talk with the robot).

As mentioned above, to create the two robot personalities, the tempo, pitch and volume of the robot's speech were manipulated based on the speech traits literature for extroversion and introversion \cite{Pittam94, Tusing00, Schmitz07, Tapus08}. Taking Furhat's standard setting as a baseline, the extrovert robot speech was given a higher pitch (+20\%), volume (+6dB) and speaking rate (+20\%), while the introvert robot speech has a lower pitch (-20\%), volume (-6dB) and speaking rate (-20\%) than the baseline. The interaction content was manipulated to reflect extrovert and introvert linguistic patterns based on the parameters provided in \cite{Neff10, Mairesse07}. Table \ref{tab:linguisticfeatures} lists some example statements from the extrovert and introvert robots.

\begin{table}[htbp]
 \caption{Extrovert and introvert statements and linguistic features applied (in square brackets)}
 \label{tab:linguisticfeatures}
    \begin{center}
    \begin{tabular}{p{3.5cm}p{3.5cm}}
    \hline
    \textbf{Extrovert Robo-Barista}&\textbf{Introvert Robo-Barista} \\
    \hline
    There are many capsules available. Why not let me help you? [self-reference] & There are many capsules available. Perhaps you might need some help? [downtoner hedges]\\
    How about I make you a Ristretto? [informal] & I can make you a Kazaar if you like? [downtoner hedges] \\
    It would be amazing if you can place a black capsule in the machine? [emphasizer hedges] & Can you put one of the dark blue capsules in the machine? \\
    Great! Keep exploring! There are many other robots to meet! & Make sure you do! There are many interesting and useful robots to meet! [claim complexity] \\
    Your order is ready! Enjoy your deliciously smelling Ristretto! Bye! [subject implicitness] & Your order is ready! Enjoy your Kazaar! Goodbye. \\
    \hline
    \end{tabular}
    \end{center}
\end{table}

\section{EXPERIMENT}
\subsection{Pilot}
\subsubsection{Procedure}
As a manipulation check and to address \textbf{R1},  we replicated prior work showing that subjects can recognise personality from speech traits and linguistic manipulation \cite{Schmitz07, Tapus08}. The pilot study employed a within-subject design. The procedure was ethically approved by Heriot-Watt University Department of Computer Science's ethics board, as is the main experiment. A total of 50 subjects took part and data from 5 subjects were excluded from analysis due to too fast and thus unrealistic and implausible completion times. Subjects were aged 18 to 62 (mean = 32.11, SD = 10.81), consisting of 34 females and 11 males. Recruitment was through the online Prolific\footnote{\hyperlink{https://www.prolific.co/}{https://www.prolific.co/}} platform. Pre-screening filters were applied to ensure that our subjects fall in the right age range (18-65) and are not in vulnerable user groups. The experiment was administered using Qualtrics\footnote{\hyperlink{https://www.qualtrics.com/uk/}{https://www.qualtrics.com/uk/}}. 

The subjects first provided digital consent followed by a confirmation that they have audio and video capabilities on their device. Next, they provided their gender and age information. Then they were asked to complete the agent personality perception experiment, by watching 2 videos (about 90 seconds each)\footnote{Introvert Robo-Barista: \hyperlink{https://youtu.be/BTk3d6dkIts}{https://youtu.be/BTk3d6dkIts}}\footnote{Extrovert Robo-Barista:\hyperlink{https://youtu.be/py3h8uLvqSg}{https://youtu.be/py3h8uLvqSg}} of a user interacting with the Robo-Barista speaking with differing speech prosody traits and linguistic features, as described in Section \ref{sec: Robo-Barista set-up}. The order of the videos was randomised to counterbalance any order effect. 

After each video, subjects were asked to rate 14 adjective pairs on a 5-point Likert scale, namely Semantic Differential items \cite{Mehrabian74}. The adjectives describe the Robo-Barista's behaviours as follows: active-passive; interested-indifferent; talkative-quiet, intelligent-stupid; predictable-unpredictable; consistent-inconsistent, fast-slow; polite-impolite; friendly-unfriendly; obedient-disobedient; diversified-boring; attentive-inattentive.  They also include those describing general usefulness: useful-useless; practical-impractical \cite{Lohse08}. As a validation check, subjects were asked to rate the robot on the introvert-extrovert dimensions on a 5-point Likert scale with 1 being introvert and 5 being extrovert. The final debrief outlined that the purpose of the survey was to assess how changes in speech prosody and linguistic features modify the perception of the robot personality on the extroversion-introversion dimensions. 

\subsubsection{Results}
Wilcoxon Signed-Rank tests were performed to assess differences across conditions. The following 7 adjective pairs showed significant differences between ratings for the two robot personalities: passive-active ($Z=-2.798, p = 0.05$), indifferent-interested ($Z=-0.256, p = 0.01$), quiet-talkative ($Z=-4.497, p < 0.001$), slow-fast ($Z=-5.065, p < 0.001$), unfriendly-friendly ($Z=-2.609, p = 0.09$), boring-diversified ($Z=-3.21, p = 0.001$) and  inattentive-attentive ($Z=-3.3, p < 0.001$).

Importantly, a significant difference was also found in the introvert-extrovert dimension rating, $Z=-4.284,p<0.001$. Therefore, the differences between the ratings of the introvert and extrovert robot were clearly perceived. This provides strong evidence that our manipulation of speech and linguistic features was successful.

\subsection{Main Experiment}
\subsubsection{Procedure}
The aim of the main experiment is to address the research questions raised in Section \ref{sec:Introduction}. 202 new subjects were recruited on Prolific. They were aged 18 to 60 (mean=28.11, SD=8.55) consisting of 77 female, 122 male and 3 non-binary. Participant recruitment and pre-screening followed the same procedure as the pilot study. 

Prior to the experiment, subjects signed a consent form and were asked to provide demographic information including age, gender and prior experiences and attitudes to robots using the 14-item NARS \cite{Nomura06} and the 6-item PTT questionnaires \cite{Jessup19}. NARS measures negative attitudes towards robots, while PTT was designed to measure stable characteristics in individuals, attitudes towards technology and whether people were likely to collaborate with technology. Both questionnaires use a 5-point Likert scale ranging from strongly disagree to strongly agree. Jessup et al. \cite{Jessup19} have shown that adapting the PTT questionnaire to the interaction context enhances the reliability of the measure, and hence the predictability of perceived trustworthiness. Following this prior-work, the PTT questionnaire was adapted to our scenario with the word ``technology" being replaced by ``robot".  We added an additional question to the NARS specific to our scenario - ``I would feel uneasy if I had to order a drink from robots". 

Then Big 5 44-item inventory \cite{McCrae04} was administered. Scores for all 5 traits were calculated for each subject and used for correlation analysis. Additionally, categorisation was performed for the extroversion trait to allow personality matching between users' and the Robo-Barista's personality. The mean extroversion score across the entire sample is M = 23.50. The recommended approach is to compare the subjects' score against a published norm or construct a local norm \cite{Srivastava12}. Given that there is no official published norm, we establish a local norm for our sample population. Assuming subjects scoring above mean as extroverts, while subjects scoring below mean as introverts, we have 98 extroverts and 103 introverts. Two subjects (score 13 and 20) have a missing value in one question each, so the one who scored 20 has been excluded from analysis, as they may fall in either dimension. 

During the experiment, subjects watched the same two short videos as in the pilot study, presented in alternating order. After each video, we repeated the pilot manipulation check and asked them to rate the videos using a 5-point Likert scale spanning the introvert (1) and extrovert (5) dimension. Then, they were asked to provide ratings of the level of their trust in the Robo-Barista using an adapted LETRAS-G \cite{Kraus20}, a subset of the empirically determined scale of trust in automated systems \cite{Jian00}. The scale contains 7 items (listed below) on a 7-point Likert ranging from not at all (1) to extremely (7), designed to measure the subject's trust in a specific robot. Thus two trust scores, one for each version of the Robo-Barista were computed for analysis.
\begin{itemize}
    \item \textbf{LG1:} I trust the robot
    \item \textbf{LG2:} The robot is reliable
    \item \textbf{LG3:} The robot is trustworthy
    \item \textbf{LG4:} I am sceptical about the robot - Reverse
    \item \textbf{LG5:} I distrust the robot’s suggestions – Reverse
    \item \textbf{LG6:} I am suspicious of the robot - Reverse
    \item \textbf{LG7:} Using the robot will lead to dangerous and harmful consequences - Reverse
\end{itemize}

The subjects were also asked to rate how much they like the robot’s voice ranging from strong dislike (1) to strong like (5) and whether they can hear the speech clearly ranging from not at all (1) to all the time (5). After watching and rating both videos, the subjects were asked to select: 1) their preferred robot and 2) the robot they trusted more. They were asked to explain their choices for preference and trust through worded qualitative feedback. 

\subsubsection{Independent and Dependent Variables}
This study has a within-subjects design with the independent variables as below. \textbf{IV1} and \textbf{IV4} are manipulated while \textbf{IV2}, \textbf{IV3} and \textbf{IV5} are non-manipulated variables.
\begin{itemize}
    \item \textbf{IV1:} The Robo-Barista personality (Introvert/Extrovert)
    \item \textbf{IV2:} The subjects' PTT score
    \item \textbf{IV3:} The subjects' NARS score (with the extra question specific to our scenario)
    \item \textbf{IV4:} The subjects' Introvert/Extrovert category
    \item \textbf{IV5:} The subjects' scores for each Big 5 trait (Big 5)
\end{itemize} 

The dependent variables are: 
\begin{itemize}
    \item \textbf{DV1:} The subjects' perception of the Robo-Barista’s personality
    \item \textbf{DV2:} The choice of Robo-Barista personality that the subjects preferred for interaction
    \item \textbf{DV3:} The choice of Robo-Barista personality that the subjects trusted more
    \item \textbf{DV4:} The subjects' levels of trust in the Extrovert and Introvert Robo-Baristas, as measured using the LETRAS-G instrument
\end{itemize}

\section{QUANTITATIVE RESULTS}
\subsection{\textbf{R1}: Perceived Robo-Barista Personality}
With regards \textbf{R1}, a Wilcoxon Signed-Rank test indicated that the Extrovert Robot was rated significantly higher on the introversion-extroversion scale than the Introvert Robot ($Z=-8.737, p < 0.001$), confirming that the speech traits and linguistic features manipulation manifested in perceivable robot personalities.

\subsection{\textbf{R2}: Robo-Barista Personality and Preference/Trust}
To answer \textbf{R2}, we performed binomial tests on subjects' choice of preferred and trusted Robo-Barista.
\subsubsection{Personality Preference} The binomial test on preference shows that an Extrovert Robot is preferred over an Introvert Robot, $p < 0.001$ (n=130 for Extrovert Robot and n=72 for Introvert Robot). 

\subsubsection{Personality Trust} A further binomial test indicates that an Extrovert Robot is trusted more than an Introvert Robot, $p < 0.001$ (n=125 for Extrovert Robot and n=77 for Introvert Robot). Paired t-tests showed significant differences in 5 LETRAS-G items and the overall trust levels (\textbf{DV4}) between the two Robo-Baristas in favour of the Extrovert Robot, as listed in Table \ref{tab:LETRASG}.
 
\begin{table}[htbp]
  \caption{Paired samples T-test for Likert scale ratings of the two Robo-Baristas' assigned trust levels using LETRAS-G items. * indicates significant. }
 \label{tab:LETRASG}
    \begin{center}
    \begin{tabular}{|l|c|c|c|c|}
    \hline
    Item & Mean \textbf{E} & Mean \textbf{I} & t & p-value \\
    \hline
    \textbf{LG1:} Trust & \textbf{4.84} & 4.64 & 2.495 & 0.013*\\
    \textbf{LG2:} Reliability & \textbf{5.22} & 4.88 & 4.257 & 0.000*\\
    \textbf{LG3:} Trustworthy & \textbf{5.00} & 4.74 & 3.018 & 0.003*\\
    \textbf{LG4:} Scepticism & 3.64 & \textbf{3.94} & -3.030 & 0.003*\\
    \textbf{LG5:} Distrust & 2.98 & 3.08 & -1.273 & 0.205\\
    \textbf{LG6:} Suspicion & 3.10 & \textbf{3.28} & -2.123 & 0.035*\\
    \textbf{LG7:} Dangerous & 2.50 & 2.45 & 0.642 & 0.522\\
    Overall & \textbf{4.98} & 4.79 & 3.804 & 0.000*\\
    \hline
    \end{tabular}
    \end{center}
\end{table} 

\subsubsection{Similarity/Complementary-Attraction}
\begin{figure}[htbp]
    \centering
    \framebox{\includegraphics[scale=0.7]{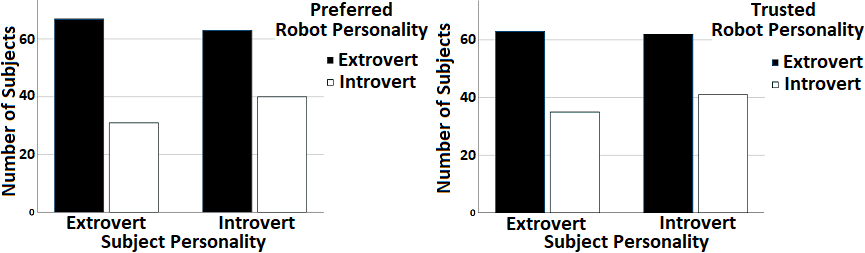}}
    \caption{``Preferred"  - Left chart (\textbf{DV2}) and ``More Trusted" - Right chart (\textbf{DV3}) Robo-Barista personality based on subjects' binary choice answers grouped by subjects' Introvert/Extrovert category (\textbf{IV4})}
    \label{fig:personalityvspreferredandtrusted}
\end{figure}

Fig. \ref{fig:personalityvspreferredandtrusted} shows preference and trust results grouped by the subject's Introvert/Extrovert category. A Chi-square test for independence showed a non-significant association between the preferred Robo-Barista personality (\textbf{DV2}) and the subjects' Introvert/Extrovert category (\textbf{IV4}), ${\chi}^2(4) = 1.14, p = 0.286$. Subjects' Introvert/Extrovert category also showed no influence on the choice of more trusted Robo-Barista personality, confirmed by a Chi-square test, ${\chi}^2(4) = 0.358, p = 0.55$.  These results do not, therefore, show evidence for the similarity/complementary-attraction theories, answering \textbf{R2}. 

\subsection{\textbf{R3}: Preference and Trust Correlation}
To address \textbf{R3}, a Chi-square test for independence was computed to determine whether the choice of Robo-Barista personality that the subjects trust more (\textbf{DV3}) relates to the choice of Robo-Barista personality the subjects preferred (\textbf{DV2}). An association was observed, ${\chi}^2(4) = 109.24, p < 0.001$. A Pearson correlation established that the choice of preferred Robo-Barista correlates positively with the choice of more trusted Robo-Barista $r(202) = 0.735, p < 0.001$.

\subsection{\textbf{R4}: Trust and Individual Differences}
This section reports results looking at a subject's Big 5 traits and characteristics, such as PTT and NARS, and exploring if these affect the way they rate the trust levels of the two variations of the robot.

\subsubsection{Trust Scores and Big 5 traits} 
We explore the link between subjects' Big 5 (\textbf{IV5}) and their trust scores (\textbf{DV4}), separated out into how they rated the Extrovert and Introvert Robot. Subjects with missing values for a trait were excluded from analysis. Whilst there was no clear signal for most of the traits, we did find some interesting results. Specifically, a Pearson correlation coefficient showed positive correlation between Extrovert Robot trust score and subject agreeableness $r(199) = 0.211, p < 0.05$. Moreover, a near significant correlation was observed between Introvert Robot trust score and subject neuroticism $r(201) = -0.135, p = 0.056$. 

\subsubsection{Trust Scores and PTT}
Next, we explore the link between subjects' PTT (\textbf{IV2}) and their trust scores (\textbf{DV4}). Subjects' PTT and trust score for the Extrovert Robot were found to be significantly positively correlated, $r(202) =  0.478, p < 0.001$. A significant positive correlation was also found between subjects' PTT and trust score for the Introvert Robot, $r(202) = 0.484, p < 0.001$. 

\subsubsection{Trust Scores and NARS} 
We also investigated if subjects' general negative attitudes to robots (\textbf{IV3}) is somehow linked to their trust scores (\textbf{DV4}). A Pearson correlation coefficient was computed for NARS and trust level in the Extrovert Robot, $r(202)=-0.554,p<0.001$; as well as NARS and trust level in the Introvert Robot, $r(202)=-0.584,p<0.001$ resulting in significant negative correlation in both cases.

\subsection{Voice and Speech of Robo-Baristas}
\label{sec: Voice and Speech}
To investigate subjects' opinion on the Robo-Baristas' voice and speech, Wilcoxon Signed-Rank tests were conducted to compare their 5-point Likert scale ratings on how much they like the Introvert and Extrovert Robots' voice and how clearly they can hear the speech. The results indicated that subjects liked the Extrovert Robot voice more than the Introvert Robot voice ($Z=-4.754, p < 0.001$), while the Introvert Robot speech was heard more clearly than the Extrovert Robot speech ($Z=-4.286, p < 0.001$). 

\subsection{Confounding Variables}
Mixed ANOVAs were conducted to explore whether there was an ordering effect between: ExtrovertIntrovert (those who watched the extrovert Robo-Barista video first) and IntrovertExtrovert (those who watched the introvert Robo-Barista video first) and the following variable: Robo-Barista personality ratings (Introvert, Extrovert); gender (male, female, non-binary) and Robo-Barista personality ratings; as well as age (10-19, 20-29, 30-39, 40-49, 50+) and Robo-Barista personality ratings. There was a statistically significant difference between ordering groups and Robo-Barista personality ratings, $F(1,200)=10.631, p=0.001$. A Bonferroni corrected post-hoc test revealed that the IntrovertExtrovert group ratings of the Robo-Barista personality were statistically significantly higher than the ratings of the ExtrovertIntrovert group, $p = 0.001$. One possible explanation might be due to them rating the Introvert Robot quite highly on the introvert-extrovert dimension at the start resulting in them giving an even higher rating for the Extrovert Robot later. No significant interaction effect was found between order groups and personality ratings. For gender and age effect, no significant main effect was found, $F(2,199)=2.072, p=0.129$ and $F(4,197)=0.816, p=0.516$ respectively. The interaction effects between order groups, gender, age and personality ratings were also non-significant, ruling them out as confounding variables in our study.

\section{QUALITATIVE RESULTS}
\label{sec: Qualitative results}
\subsection{Codebook}
Qualitative analysis was carried out on the reasons subjects provided for their robot preference and for trusting more in a specific Robo-Barista. A codebook was developed by two senior researchers, using an inductive approach similar to that used in grounded theory \cite{McDonald19}. The initial codebook was then tested by the two researchers independently coding 10\% of the questionnaires and disagreements were discussed. 
A team of 2 coders was then established and a small sample of questionnaires were coded individually, disagreements were resolved and clarifications were added to the codebook. Finally, a random sample of 30\% of the questionnaires were dual coded independently. The average agreement across all codes was 99\% with a Cohen’s Kappa of 0.93. 

The codebook contains codes referring to the Robo-Baristas' attributes that gave rise to subjects' preference and trust (full list in Table \ref{tab:preferenceandtrustreasons}). Further codes pertain to the sense and sentiment of these attributes (i.e. if the subject saw an attribute as good or bad, whether the focus was towards the preferred robot or away from that not preferred, and the valence of the attribute e.g. pleasant/unpleasant). Lastly, indifference expression where subjects were impartial about the Robo-Baristas was also coded.

\begin{table}[htbp]
 \caption{Percentage of all subjects indicating certain attributes as their preference and trust reasons}
 \label{tab:preferenceandtrustreasons}
    \begin{center}
    \begin{tabular}{|c|c|c|}
    \hline
    Attribute & Preference & Trust \\
    \hline
    Appeal & 22.3 & 12.4\\
    Better & 6.9 & 4\\
    Competence & 3 & 13.9\\
    Ease of Communication & 9.9 & 11.4\\
    Holistic View of Demeanour & 4 & 6.9 \\
    Mood & 29.2 & 28.2\\
    Subjects Affected or Moved & 3 & 9.9\\
    Tempo of Speech & 11.4 & 5.4\\
    Trustworthiness & 1.5 & 8 \\
    Understandability & 21.3 & 11.9\\
    Voice & 77.7 & 60.4\\
    \hline
    \end{tabular}
    \end{center}
\end{table} 

\subsection{Indifference Expression}
It can be observed that all subjects (n=202) expressed a reason for preference with Voice being the main determinant (77.7\%). When it comes to trust, 8.4\% expressed no inclination towards either Robo-Barista stating that they trusted both equally. Voice again was the main factor for trust (60.4\%). Voice was the main manipulated attribute differentiating both Robo-Baristas, so this finding was as expected.

\subsection{Attributes Reference}
Next, we looked at Holistic View of Demeanour of the Robo-Baristas, indicating that the subject was referring to something other than only the voice. 4\% of all subjects indicated this as their preference reason while 6.9\% indicated this as their trust reason. e.g.  \textit{“It seemed more thoughtful”} [P185] referring to the Introvert Robot, which speaks slower and with lower pitch, \textit{“Looks like a normal person being polite”} [P91] referring to the Extrovert Robot.

Looking at Appeal (enjoyable, likeability, pleasantness/niceness), Better (something unspecified about the robot is better) and Understandability of the conversations, more subjects indicated these as their preference reason than trust reason. This is mainly due to the Robo-Baristas' voice and speech, e.g.  \textit{“It speaks in a more natural cadence...more enjoyable”} [P82], \textit{“Clear speech, pleasant voice”} [P155]. 

In terms of Ease of communication, (i.e. how comfortable the subjects are at communicating with the Robo-Baristas) and Mood, (i.e. energy, friendliness, kindness, relaxed, warmness of the Robo-Baristas), the mentions for preference reason and trust reason are balanced. 

\subsection{Trust-Related Attributes}
Competence of the Robo-Barista was deemed more important for trust than preference (3\% for preference and 13.9\% for trust reason). The perception of the Robo-Barista credibility seemed to be at play here, e.g. \textit{“Convincing”} [P20], \textit{“...sounds more confident...”} [P43].

This is directly linked to Trustworthiness, which was also more crucial as a trust reason than a preference reason. 8\% of subjects addressed directly, aspects of Trustworthiness, such as the Robo-Barista being more genuine, reliable or explicitly trustworthy in their trust reason but only 1.5\% mention this in the preference reason. The two subthemes here were: a) those inspired to trust more the Introvert Robot due to its calm delivery and seeming reliability e.g. \textit{“By speaking a little slower, it creates an atmosphere of trust”} [P149]; and b) those trusting more the Extrovert Robot, due to its cheerful clarity perhaps reflecting what they would expect from a Barista e.g. \textit{“... more reliable and more normal to the standards.”} [P102], \textit{“ ...upbeat attitude...”}[P46]. This perception of slow speech tempo as reliable, and fast as more normal for a Barista and thus more trustworthy, is also reflected when subjects commented on the Tempo of Speech. 5\% preferred and 3.5\% trusted more the slower talking Introvert Robot, 
while 3\% preferred and 2\% trusted more the faster speaking Extrovert Robot.

Another focal point for trust is how the subjects were Affected or Moved by at least one of the Robo-Baristas. 3\% indicated this in their preference reason e.g. \textit{``...it brings me peace of mind"} [P149] and \textit{``... gives some creepy feelings"} [P24]. In contrast, 10\% indicated this in their trust reason e.g. \textit{``...made me feel more confident..."} [P37] and \textit{``Made me feel more relaxed"} [P194]. 
Thus, we can see that more subjects described feeling affected in relation to their trust reason than their preference reason. This may indicate that, for many, trust was an affective feeling as opposed to a cognitive conclusion, reflecting the theoretical categorisation of trust into cognitive and affective trust \cite{johnson2005cognitive}. Cognitive trust (someone's readiness to rely on a service due to competence and reliability) is based on knowledge while affective trust (someone's confidence in a service provider based on the care and concern they demonstrate) is emotional in nature.

\section{DISCUSSION}
\label{sec: Discussion}
From the results, we can conclude that subjects were able to distinguish the two different Robo-Barista personalities, realised through voice and lexicon manipulation. Subjects were found to prefer an Extrovert Robot over an Introvert Robot and they also trusted the Extrovert Robot more than the Introvert Robot. This results in a positive relationship between subjects' preferred Robo-Barista personality and the Robo-Barista personality they trusted more. The results validate the conclusion deduced from \cite{Howarth69, Sandstrom14} that extrovert Baristas will be preferred due to prior stereotype and they are deemed to be more likely to engage in small talk enabling customers to treat them like a ``weak social tie" \cite{Sandstrom14}. 
    
Furthermore, all LETRAS-G trust items except for two showed significant differences favouring the Extrovert Robot, consistent with the finding above. No significant differences were found for \textbf{LG5:}Distrust and \textbf{LG7:}Dangerous, which is perhaps not surprising as both robots performed their task successfully, giving no reason for subjects to distrust them and they did not portray any harmful behaviour. This might also explain why some subjects expressed indifference in their subjective responses as to which Robo-Barista they trusted more. The insignificant results of \textbf{LG5:}Distrust and \textbf{LG7:}Dangerous suggest that these items are not necessarily important to our scenario, concurring with \cite{Chita-Tegmark21} that not all items in existing trust scales are always applicable but instead are context-dependent.

Interestingly, subjects' own Introvert/Extrovert personality category did not show any influence on the Robo-Barista personality they preferred or trusted more. Both introverts and extroverts preferred and trusted the Extrovert Robot more, as illustrated in Fig. \ref{fig:personalityvspreferredandtrusted}. Although this result contradicts the personality matching theories \cite{Whittaker21, Lee06}, it is inline with Joosse et al.'s \cite{Joosse13} finding that task constraints add complexity to human-robot trust relationship. One explanation could be that getting coffee is a social activity, hence the outgoing attributes of an extrovert are more favourable in general, as it conforms to existing stereotype \cite{Howarth69}. 

There was no relation observed between people's predisposition to trust a robot and their own personality (Extrovert/Introvert), contradicting \cite{Merritt08, Evans08} but confirming Alarcon et al.'s \cite{Alarcon18} findings. However, we did find an effect that people who are more conscientious have lower PTT, $r(199) = -0.158, p < 0.05$, thus are generally less likely to trust. In addition, people who are more agreeable showed higher perceived trustworthiness especially in the Extrovert Robot, which they trusted more. There was also a trend that people who are more neurotic had lower perceived trustworthiness in the Introvert Robot, which they trusted less. These relationships between Big 5 traits and PTT, as well as, perceived trustworthiness conform to prior work in psychology and behavioural economics \cite{Evans08, Becker12, Calhoun17, Dohmen08}.

Perhaps unsurprising, we found that subjects' PTT in robots positively predicts their trust in the Robo-Barista. This is the case for both the Introvert and Extrovert Robots. Additionally, their prior NARS negatively predict their trust in the Robo-Baristas. As a result, subjects' PTT in robots is also negatively related to their NARS, $r(202) =  -0.575, p < 0.001$. This indicates that generally those with higher positive attitudes to robots tend to have a higher propensity to trust them. This result signifies the importance of gathering information on users' PTT and prior attitudes in robots when investigating human-robot trust, as this can help in understanding the processes underlying trust interactions, and hence, inform appropriate robot design to context \cite{Lewis18}.

In general as observable from Section \ref{sec: Voice and Speech}, subjects preferred the Extrovert Robot voice although the Introvert Robot speech was found to be clearer. Voice was also the main determinant for preference and trust. This is mainly due to the cheerful, friendly and appealing traits of the Extrovert Robot's voice and calm delivery and clarity of the Introvert Robot's speech.  Both these qualities inspired trust, suggesting the importance of voice characteristics in moderating user trust. Additionally, the perceived competency, trustworthiness and affective influence of the robot are other factors affecting trust perception.

\section{CONCLUSIONS AND FUTURE WORK}
This paper investigates how users' individual differences affect trust in the Robo-Baristas. Although providing the first step to investigate HRI for the service industry, it has some limitations that provide future research opportunities. 

Our study revealed that subjects' predispositions (i.e. propensity to trust in robots and negative attitudes towards robots) predicted their levels of trust in the Robo-Baristas. Furthermore, to a certain extent subjects' Big 5 personality traits influenced their PTT and perceived trustworthiness of the Robo-Baristas. The main implication is that any human-robot trust study should consider collecting this information as it can inform context-appropriate robot design. Overall, subjects preferred and trusted an extrovert Robo-Barista more than an introvert Robo-Barista, irrespective of their own personality, suggesting the importance of context as well as prior stereotypes. Another implication of our findings is that voice features should be taken account when designing social robots and the resulting voice should sound trustworthy relative to the robot's role. Impression of competence and trustworthiness are other factors to consider to prevent misalignment of users' expectations and robot capabilities, which might lead to trust breakdown.

It would be interesting in future work to investigate not only subjects' personality but subjects' stereotypical thoughts on the task and the associated personality. Furthermore, instead of observing video vignettes of the interactions, in-person human-robot interactions where subjects order drinks from the Robo-Baristas is a direction worth exploring, as the relationship of personality with trust and performance in HRI may be moderated by the type of study \cite{Alarcon21}. 
Given that mental models of robots are context-sensitive, in future subjects should be given an option to flag out non-applicable items in the trust scale used \cite{Chita-Tegmark21} and these ratings should be considered when reporting and interpreting results. Cultural factors also play a role in how individuals trust robotic agents \cite{Lewis18}, providing another interesting research avenue. Finally, with trust being a dynamic process \cite{Lee04} and social robots being built to support long-term interactions, we are currently planning a long-term ``in the wild" study to investigate trust formation and development over time in the caf\'e robot context.

\bibliographystyle{IEEEtran} 
\bibliography{IEEEabrv,Robotcafe}

\begin{thebibliography}{10}
\providecommand{\url}[1]{#1}
\csname url@samestyle\endcsname
\providecommand{\newblock}{\relax}
\providecommand{\bibinfo}[2]{#2}
\providecommand{\BIBentrySTDinterwordspacing}{\spaceskip=0pt\relax}
\providecommand{\BIBentryALTinterwordstretchfactor}{4}
\providecommand{\BIBentryALTinterwordspacing}{\spaceskip=\fontdimen2\font plus
\BIBentryALTinterwordstretchfactor\fontdimen3\font minus
  \fontdimen4\font\relax}
\providecommand{\BIBforeignlanguage}[2]{{%
\expandafter\ifx\csname l@#1\endcsname\relax
\typeout{** WARNING: IEEEtran.bst: No hyphenation pattern has been}%
\typeout{** loaded for the language `#1'. Using the pattern for}%
\typeout{** the default language instead.}%
\else
\language=\csname l@#1\endcsname
\fi
#2}}
\providecommand{\BIBdecl}{\relax}
\BIBdecl

\bibitem{Lim22}
M.~Y. Lim, J.~David Aguas~Lopes, D.~A. Robb, B.~W. Wilson, M.~M., and
  H.~Hastie, ``Demonstration of a robo-barista for in the wild interactions,''
  in \emph{Proc. ACM/IEEE International Conference on Human-Robot Interaction
  (HRI’22)}, Sapporo, Hokaido, Japan, 2022.

\bibitem{Meriam22}
O.~L. Meriam~Moujahid, Helen~Hastie, ``Multi-party interaction with a robot
  receptionist,'' in \emph{Proc. of the ACM/IEEE International Conference on
  Human-Robot Interaction}, 2022.

\bibitem{Keizer14}
S.~Keizer, M.~E. Foster, Z.~Wang, and O.~Lemon, ``Machine learning for social
  multiparty human--robot interaction,'' \emph{ACM Trans. on Interactive
  Intell. Syst.}, vol.~4, no.~14, pp. 1--32, 2014.

\bibitem{Wynne18}
K.~T. Wynne and J.~B. Lyons, ``An integrative model of autonomous agent
  teammate likeness,'' \emph{Theor. Issues in Ergonom. Sci.}, vol.~19, pp.
  353--374, 2018.

\bibitem{Hancock21}
P.~A. Hancock, T.~T. Kessler, A.~D. Kaplan, J.~C. Brill, and J.~L. Szalma,
  ``Evolving trust in robots: Specification through sequential and comparative
  meta-analyses,'' \emph{Human factors}, vol.~63, no.~7, pp. 1196--1229, 2021.

\bibitem{Joosse13}
M.~Joosse, M.~Lohse, J.~G. Pérez, and V.~Evers, ``What you do is who you are:
  The role of task context in perceived social robot personality,'' in
  \emph{Proc. of Robotics and automation 2013 IEEE international
  conference}.\hskip 1em plus 0.5em minus 0.4em\relax Karlsruhe, Germany: IEEE,
  May 2013, pp. 2134--2139.

\bibitem{Howarth69}
E.~Howarth, ``Expectations concerning occupations in relation to
  extraversion-introversion,'' \emph{Psychological Reports}, vol.~24, no.~2,
  pp. 415--418, Nov. 1999.

\bibitem{Sandstrom14}
G.~M. Sandstrom and E.~W. Dunn, ``Is efficiency overrated?: Minimal social
  interactions lead to belonging and positive affect,'' \emph{Social
  Psychological and Personality Sci.}, vol.~5, no.~4, pp. 437--442, 2014.

\bibitem{Lee04}
J.~D. Lee and K.~A. See, ``Trust in automation: Designing for appropriate
  reliance,'' \emph{Human Factors: The J. of the Human Factors and Ergonom.
  Soc.}, vol.~46, pp. 50--80, 2004.

\bibitem{Freedy07}
A.~Freedy, E.~DeVisser, G.~Weltman, and N.~Coeyman, ``Measurement of trust in
  human–robot collaboration,'' in \emph{Proc. of International symposium on
  collaborative technologies and systems}, 2007, pp. 106--114.

\bibitem{Kraus20}
J.~M. Kraus, ``Psychological processes in the formation and calibration of
  trust in automation,'' Ph.D. dissertation, Ulm University, Germany, 2020.

\bibitem{Nomura06}
T.~Nomura, T.~Suzuki, T.~Kanda, and K.~Kato, ``Measurement of negative
  attitudes toward robots,'' \emph{Interact. Studies: Social Behav. and Commun.
  in Biol. and Artif. Syst.}, vol.~7, no.~3, pp. 437--454, 2006.

\bibitem{Mayer95}
R.~C. Mayer, J.~H. Davis, and F.~D. Schoorman, ``An integrative model of
  organizational trust,'' \emph{Acad. of Manage. Rev.}, vol.~20, pp. 709--734,
  1995.

\bibitem{McKnight98}
D.~H. McKnight, L.~L. Cummings, and N.~L. Chervany, ``Initial trust formation
  in new organizational relationships,'' \emph{Acad. of Manage. Rev.}, vol.~23,
  no.~3, pp. 473--490, 1998.

\bibitem{Rotter67}
J.~B. Rotter, ``A new scale for the measurement of interpersonal trust,''
  \emph{J. of personality}, vol.~35, no.~4, pp. 651--665, 1967.

\bibitem{Jessup19}
S.~A. Jessup, T.~Schneider, G.~Alarcon, T.~Ryan, and A.~Capiola, ``The
  measurement of the propensity to trust automation,'' in \emph{Virtual,
  Augmented and Mixed Reality. Applications and Case Studies. HCII 2019.
  Lecture Notes in Computer Science}, C.~J. and F.~G., Eds.\hskip 1em plus
  0.5em minus 0.4em\relax Cham: Springer, 2019, vol. 11575.

\bibitem{Lewis18}
M.~Lewis, K.~Sycara, and P.~Walker, ``The role of trust in human-robot
  interaction.'' \emph{Found. of Trusted Autonomy}, 2018.

\bibitem{McCrae04}
R.~R. McCrae, ``Human nature and culture: A trait perspective,'' \emph{J. of
  Res. in Personality}, vol.~38, pp. 3--14, 2004.

\bibitem{Gaines97}
S.~O. Gaines, A.~T. Panter, M.~D. Lyde, W.~N. Steers, C.~E. Rusbult, C.~L. Cox,
  and M.~O. Wexler, ``Evaluating the circumplexity of interpersonal traits and
  the manifestation of interpersonal traits in interpersonal trust.'' \emph{J.
  of Personality and Social Psychol.}, vol.~73, no.~3, pp. 610--623, 1997.

\bibitem{Merritt08}
S.~M. Merritt and D.~R. Ilgen, ``Not all trust is created equal: Dispositional
  and history-based trust in human-automation interactions,'' \emph{Human
  Factors}, vol.~50, no.~2, pp. 194--210, 2008.

\bibitem{Evans08}
A.~M. Evans and W.~Revelle, ``Survey and behavioral measurements of
  interpersonal trust,'' \emph{J. of Res. in Personality}, vol.~42, p.
  1585–1593, 2008.

\bibitem{Robert18}
L.~Robert, ``Personality in the human robot interaction literature: A review
  and brief critique,'' in \emph{Proc. of the 24th Americas conference on
  information systems}, L.~P. Robert, Ed., New Orleans, LA, Aug. 2018, pp.
  16--18.

\bibitem{Schmitz07}
M.~Schmitz, A.~Krüger, and S.~Schmidt, ``Modeling personality in voice of
  talking products through prosodic parameters,'' in \emph{Proc. of IUI 2007–
  12th International Conference on Intelligent User Interfaces}, Honolulu,
  Hawaii, USA, Jan 2007, pp. 313--316.

\bibitem{Tapus08}
A.~Tapus, C.~Ţăpuş, , and M.~Mataric, ``User-robot personality matching and
  assistive robot behavior adaptation for post-stroke rehabilitation therapy,''
  \emph{Intell. Service Robot.}, vol.~1, no.~2, pp. 169--183, 2008.

\bibitem{Pittam94}
J.~Pittam, \emph{Voice in social interaction: An interdisciplinary
  approach}.\hskip 1em plus 0.5em minus 0.4em\relax Thousand Oaks, CA: Sage,
  1994.

\bibitem{Tusing00}
K.~J. Tusing and J.~P. Dillard, ``The sounds of dominance: Vocal precursors of
  perceived dominance during interpersonal influence,'' \emph{Human Commun.
  Res.}, vol.~26, no.~1, pp. 148--171, 2000.

\bibitem{Mairesse07}
F.~Mairesse and M.~Walker, ``{PERSONAGE}: Personality generation for
  dialogue,'' in \emph{Proc. of the 45th Annual Meeting of the Association of
  Computational Linguistics}.\hskip 1em plus 0.5em minus 0.4em\relax Prague,
  Czech Republic: Association for Computational Linguistics, Jun. 2007, pp.
  496--503.

\bibitem{Neff10}
M.~Neff, Y.~Wang, R.~Abbott, and M.~Walker, ``Evaluating the effect of gesture
  and language on personality perception in conversational agents,'' in
  \emph{Proc. IVA 2010, Lecture Notes in Computer Science}, J.~Allbeck,
  N.~Badler, T.~Bickmore, C.~Pelachaud, and A.~Safonova, Eds.\hskip 1em plus
  0.5em minus 0.4em\relax Berlin, Heidelberg: Springer, 2010.

\bibitem{Furham90}
A.~Furnham, \emph{Handbook of Language and Social Psychology}.\hskip 1em plus
  0.5em minus 0.4em\relax Winley, 1990.

\bibitem{Lohse08}
M.~L. et~al., ``Evaluating extrovert and introvert behaviour of a domestic
  robot — a video study,'' in \emph{Proc. of The 17th IEEE International
  Symposium on Robot and Human Interactive Communication}, Munich, 2008, pp.
  488--493.

\bibitem{Leuwerink12}
K.~Leuwerink, ``A robot with personality: Interacting with a group of humans,''
  in \emph{Proc. of the 16th Twente Student Conference on IT}, vol.~4,
  Enschede, The Netherlands, 2012.

\bibitem{Alarcon18}
G.~Alarcon, J.~Lyons, J.~Christensen, M.~Bowers, S.~Klosterman, and A.~Capiola,
  ``The role of propensity to trust and the five factor model across the trust
  process,'' \emph{J. Res. Pers.}, vol.~75, pp. 69--82, 2018.

\bibitem{Whittaker21}
S.~Whittaker, Y.~Rogers, E.~Petrovskaya, and H.~Zhuang, ``Designing personas
  for expressive robots: Personality in the new breed of moving, speaking, and
  colorful social home robots,'' \emph{ACM Trans. Human-Robot Interact.},
  vol.~10, no.~8, p. 25 pages, February 2021.

\bibitem{Lee06}
K.~M. Lee, W.~Peng, S.-A. Jin, and C.~Yan, ``Can robots manifest personality?:
  An empirical test of personality recognition, social responses, and social
  presence in human–robot interaction,'' \emph{J. of Commun.}, vol.~56, pp.
  754--772, 2006.

\bibitem{Rousseau98}
D.~M. Rousseau, S.~B. Sitkin, R.~S. Burt, and C.~Camerer, ``Not so different
  after all: A cross-discipline view of trust,'' \emph{Acad. of Manage. Rev.},
  vol.~23, no.~3, pp. 393--404, 1998.

\bibitem{Milleounis15}
A.~Mileounis, R.~H. Cuijpers, and E.~I. Barakova, ``Creating robots with
  personality: The effect of personality on social intelligence.'' in
  \emph{Artificial Computation in Biology and Medicine. IWINAC 2015. Lecture
  Notes in Computer Science}, F.~V. J., Álvarez Sánchez~J., de~la Paz
  López~F., T.-M. F., and A.~H., Eds.\hskip 1em plus 0.5em minus 0.4em\relax
  Cham: Springer, 2015, vol. 9107.

\bibitem{Lopes21}
J.~Lopes and H.~Hastie, ``The language of persuasion, negotiation and trust,''
  in \emph{Proc. of Semdial}, 2021.

\bibitem{Rheu21}
M.~Rheu, J.~Y. Shin, W.~Peng, and J.~Huh-Yoo, ``Systematic review:
  Trust-building factors and implications for conversational agent design,''
  \emph{Int. J. of Human–Computer Interact.}, vol.~37, no.~1, pp. 81--96,
  2021.

\bibitem{bunk2020diet}
T.~Bunk, D.~Varshneya, V.~Vlasov, and A.~Nichol, ``Diet: Lightweight language
  understanding for dialogue systems,'' \emph{arXiv preprint arXiv:2004.09936},
  2020.

\bibitem{Bert19}
J.~Devlin, M.-W. Chang, K.~Lee, and K.~Toutanova, ``{BERT}: Pre-training of
  deep bidirectional transformers for language understanding,'' in \emph{Proc.
  of NAACL}, Jun. 2019, pp. 4171--4186.

\bibitem{vlasov2019dialogue}
V.~Vlasov, J.~E. Mosig, and A.~Nichol, ``Dialogue transformers,'' \emph{arXiv
  preprint arXiv:1910.00486}, 2019.

\bibitem{Mehrabian74}
A.~Mehrabian and J.~A. Russell, \emph{An approach to environmental
  psychology.}\hskip 1em plus 0.5em minus 0.4em\relax the MIT Press, 1974.

\bibitem{Srivastava12}
\BIBentryALTinterwordspacing
S.~Srivastava. (2012) Norms for the big five inventory and other personality
  measures. [Online]. Available:
  \url{https://thehardestscience.com/2012/10/17/norms-for-the-big-five-inventory-and-other-personality-measures/}
\BIBentrySTDinterwordspacing

\bibitem{Jian00}
J.-Y. Jian, A.~Bisantz, and C.~Drury, ``Foundations for an empirically
  determined scale of trust in automated systems,'' \emph{Int. J. of Cogn.
  Ergonom.}, vol.~4, pp. 53--71, 2000.

\bibitem{McDonald19}
N.~McDonald, S.~Schoenebeck, and A.~Forte., ``Reliability and inter-rater
  reliability in qualitative research: Norms and guidelines for cscw and hci
  practice,'' in \emph{Proc. ACM Human Computer Interaction}, no.~39, November
  2019.

\bibitem{johnson2005cognitive}
D.~Johnson and K.~Grayson, ``Cognitive and affective trust in service
  relationships,'' \emph{J. of Bus. Res.}, vol.~58, no.~4, pp. 500--507, 2005.

\bibitem{Chita-Tegmark21}
T.~Law, N.~Rabb, and M.~Scheutz, ``Can you trust your trust measure?'' in
  \emph{Proc. of the 2021 ACM/IEEE International Conference on Human-Robot
  Interaction}.\hskip 1em plus 0.5em minus 0.4em\relax New York, USA: ACM,
  March 2021, p. 9 pages.

\bibitem{Becker12}
A.~Becker, T.~Deckers, T.~Dohmen, A.~Falk, and F.~Kosse, ``The relationship
  between economic preferences and psychological personality measures,''
  \emph{Annual Review of Econ.}, vol.~4, pp. 453--478, 2012.

\bibitem{Calhoun17}
C.~Calhoun, P.~Bobko, M.~Schuelke, S.~Jessup, T.~Ryan, C.~Walter,
  L.~Hirshfield, N.~Bowling, C.~Bragg, S.~Khazon, A.~Nelson, G.~Alarcon, and
  C.~Stokes, ``Suspicion, trust, and automation,'' Tech. Rep. SRA Int. Inc.
  Publication No. AFRL-RH-WP-TR-2017-0002, 2017.

\bibitem{Dohmen08}
T.~Dohmen, A.~Falk, D.~Huffman, and U.~Sunde, ``Representative trust and
  reciprocity: Prevalence and determinants,'' \emph{Econ. Inquiry}, vol.~46,
  pp. 84--90, 2008.

\bibitem{Alarcon21}
G.~M. Alarcon, A.~Capiola, and M.~D. Pfahler, ``Chapter 7 - the role of human
  personality on trust in human-robot interaction,'' in \emph{Trust in
  Human-Robot Interaction}, J.~B.~L. Chang S.~Nam, Ed.\hskip 1em plus 0.5em
  minus 0.4em\relax Academic Press, 2021, pp. 159--178.

\end{thebibliography}

\end{document}